\documentclass[11pt,a4paper]{article}
\usepackage[hyperref]{acl2020}
\usepackage{times}
\usepackage{tcolorbox}
\usepackage{makecell}
\usepackage{latexsym}
\usepackage{graphicx}
\graphicspath{ {./images/} }

\usepackage{microtype}

\aclfinalcopy 

\title{Automated Detection of Cyberbullying Against Women and Immigrants and Cross-domain Adaptability}

   \author{Thushari Atapattu$^1$,  Mahen Herath$^2$, Georgia Zhang$^1$ \and Katrina Falkner$^1$ \\ $^1$School of Computer Science, The University of Adelaide, Adelaide, Australia \\ $^2$Department of Computer Science \& Engineering, University of Moratuwa, Katubedda, Sri Lanka \\ email: thushari.atapattu@adelaide.edu.au}

\date{}

\begin{document}
\maketitle
\begin{abstract}
Cyberbullying is a prevalent and growing social problem due to the surge of social media technology usage. 
Minorities, women, and adolescents are among the common victims of cyberbullying. Despite the advancement of NLP technologies, the automated cyberbullying detection remains challenging. This paper focuses on advancing the technology using state-of-the-art NLP techniques. We use a Twitter dataset from SemEval 2019 - Task 5 (HatEval) on hate speech against \textit{women} and \textit{immigrants}. Our best performing ensemble model based on DistilBERT has achieved 0.73 and 0.74 of F1 score in the task of classifying hate speech (Task A) and aggressiveness and target (Task B) respectively. We adapt the ensemble model developed for Task A to classify offensive language in external datasets and achieved $\sim$0.7 of F1 score using three benchmark datasets, enabling promising results for cross-domain adaptability. We conduct a qualitative analysis of misclassified tweets to provide insightful recommendations for future cyberbullying research. 
\end{abstract}

\section{Introduction and Motivation}
Cyberbullying is \textit{"the repetitive use of aggressive language amongst peers with the intention to harm others through digital media"} \cite{1_Rosa2019}. Due to the surge of social media technology use, cyberbullying has become a prevalent and growing social problem. 
Unlike in the physical environment, cyberspace, in particular, online social platforms are not yet evolved sufficiently to prevent people from communicating without disclosing identities, spreading rumours, and harassing others. The risk of and potential consequences caused by cyberbullying are critical including both physical and mental health risk to victims. The impact and consequences are common to all generations (e.g. young, elderly) including emotional and psychological distress, decline in personal/academic development, anti-social behaviour, and, potentially, suicide. 

The criticality of this societal problem is demonstrated from a study by Yale University, commenting \textit{“cyberbullying victims are 2 to 9 times more likely to consider committing suicide”} across the globe.\footnote{https://theorganicagency.com/blog/life-death-consequences-cyber-bullying/} Within Australia, the eSafety Commissioner comments \textit{“one in every five Australian children aged eight to seventeen are victims of cyberbullying (2018)”}.\footnote{https://www.theguardian.com/society/2018/oct/03/one-in-five-australian-children-are-victims-ofcyberbullying-e-safety-commissioner-says} Adolescents, minorities (e.g. refugees, LGBTQI) and women are among common targets of cyberbullying. According to Bullying Statistics\footnote{http://www.bullyingstatistics.org/content/cyber-bullying-statistics.html}, over half of adolescents are victims of cyberbullying and about the same percentage are involved in bullying.

Despite recent research advancement in \textit{hate speech} detection \cite{7_Fortuna2018}, automated identification of \textit{cyberbullying} attempts (i.e. repetitive hate speech against an individual or a group) remains as a challenging subtask of NLP. Due to diverse variants of language (e.g. hate, intimidation, sarcasm, metaphors) used by bullies and the evolution of language (e.g. slang), particularly among adolescents, the automated detection of cyberbullying is extremely challenging. The example below appears to be misogynistic as it includes the term \textit{'b***h'}; however, it is manually classified as not misogyny since the slang \textit{'gay a*s b***h'} is commonly used for a male or gay person.

\begin{quote}
    “\textit{you a \textbf{gay a*s b***h} who seeks attention, STOP! I knew ever since you gonna switch up on me... I guess you did F***ING SNAKE A*S H*E!”}
\end{quote}


To mitigate the research and social problem of cyberbullying,
this paper focuses on advancing the technology to classify cyberbullying using state-of-the-art NLP techniques. As a case study, we focus on cyberbulling against women and immigrants. Accordingly, our first research question (RQ1) asks, \textit{Can we build machine learning models to outperform current cyberbullying classification systems on women and immigrants?}. The findings of RQ1 will lead us to explore the limitations of our models and explanations for misclassification. Hence, our second research question (RQ2) investigates, \textit{What is the content of misclassified tweets and how can we categorise them?}. Finally, to evaluate the validity of our models across external cyberbullying/hate speech datasets, our third research question (RQ3) investigates, \textit{Can we successfully validate machine learning models developed for cyberbullying detection within the context of women and immigrants for other benchmark datasets?}.

To answer our research questions, we utilise a Twitter dataset developed for SemEval 2019 - Task 5 (HatEval) \cite{4_basile2019} that includes labels for three sub tasks: 1) hate speech, 2) aggressiveness, and 3) target (individual or group). We adopt a mixed-method study, using a combination of the building of machine learning models (RQ1 \& RQ3) and qualitative content analysis (RQ2) as our methodology. We make the following main contributions:
\begin{itemize}
\item We developed and evaluated cyberbullying classification models using state-of-the-art NLP technology. Even though our model performance on Task A is either equal or slightly lower than baselines, we outperformed all previous best systems and baselines on Task B. Therefore, our ensemble models based on DistilBERT \cite{16_Sanh2019} serves as the best system as yet to classify aggressiveness and target (Task B).
\item We conducted a qualitative study to categorise misclassified tweets into meaningful codes. We distinguished six categories: lack of context (CNTX), gender-related issues (GEND), issue with resolving slangs (SLNG), issues in the original annotation (ERROR), misclassified by our model (MSCL), and issues not belong to any category (OTHER) emerged from our data, establishing a point of reference for future researchers in cyberbullying, particularly, within the context of minorities (e.g. women, LGBTQI, immigrants).
\item We adopted our best pre-trained model to evaluate other benchmark datasets, including OffensEval challenge \cite{5_zampieri2019,17_zampieri2020} and Hate \& Offense task \cite{6_Davidson2017}. Our model generalised reasonably well ($\sim$0.7) with both tasks, contributing to developing a generalised model across different cyberbullying-related tasks.
\end{itemize}

\section{Background and Related Work}

Cyberbullying is a complex phenomenon that needs multiple psychological, linguistic, and social theories to understand its nature. The identification of cyberbullying is inherently more complex even for humans (except victims) as it involves repetitive behaviour, peer-oriented nature, and intentionality to harm. Therefore, we utilise a definition stated in a recent systematic literature review on cyberbullying \cite{1_Rosa2019} as \textit{“repetitive use of aggressive language amongst peers with the intention to harm others through digital media”}.

Some recent studies \cite{7_Fortuna2018} including WOAH\footnote{https://www.workshopononlineabuse.com/home} (previously known as ALW) workshop \cite{8_Robers2019} have focused on \textit{hate speech} detection as a more general field. Despite recent advancement in \textit{hate speech detection}, recognising cyberbullying in everyday problems is primarily manual based on victim reports or manual moderation. Recent studies rely on contextual features such as demography, social network, and sentiments/emotions as features to train cyberbullying classifiers \cite{15_Charzakou2019}. 

Conversely, some related workshops such as TRAC \cite{9_kumar2018} and challenges such as HatEval \cite{4_basile2019}, OffensEval \cite{5_zampieri2019,17_zampieri2020} contributed to advance the research field by developing systems using cutting-edge NLP techniques like Universal Encoder - Fermi \cite{2_indurthi2019}, LT3 \cite{23_LT3}, ensemble of deep learning models like OpenAI's GPT and Transformer models (Team NLPR@SAPOL \cite{3_Seganti2019}), and BERT (NULI \cite{10_Liu2019}). Some of these systems have surpassed baselines and earned recognition as the best-performing systems in specific subtasks (e.g. NULI achieved 0.82 of F1 score and ranked 1st place in subtask A to classify offensive language while it ranked only in 18th place for subtask C to classify targets such as individuals, group).

Despite the promise of current systems, these models are not consistent enough to perform reasonably well within all sub tasks of cyberbullying (i.e. hate speech, aggressiveness and target). Additionally, these models were not validated across other cyberbullying-related tasks to ensure generalisability. Related literature also lacks comprehensive contributions to draw implications on why machine learning models fail to improve further. Our work focuses on addressing these three drawbacks.

\section{Research Methodology}
\textbf{Research Questions.} Our research is guided by three research questions, 
\begin{itemize}
\item \textbf{RQ1:} Can we build machine learning models to outperform current cyberbullying classification systems?
\item \textbf{RQ2:} What is the content of misclassified tweets, and how can we categorise them?
\item \textbf{RQ3:} Can we successfully validate machine learning models developed for cyberbullying detection within the context of women and immigrants for other benchmark datasets?
\end{itemize}

\noindent\textbf{Dataset.} We utilise a dataset collected from Twitter during July to September 2018 for SemEval 2019 - Task 5 (HatEval) challenge \cite{4_basile2019}. This challenge was organised to advance the technology to classify cyberbullying against women and immigrants. Tweets were collected both from English and Spanish language. We utilise only the English dataset in this paper. The dataset contains a set of tweets and their labels; HS - Hate Speech (0 - No, 1 - Yes), TR - Target Range (0 - generic group, 1 - individual), AG - Aggressiveness (0 - No, 1 - Yes). The challenge was divided into two subtasks, Task A - classification of HS, and Task B - classification of AG and TR. The dataset was labelled via AllCloud crowdsourcing platform and added two more experienced annotators to determine the final labels. Inter-rater reliability for HS, TR, and AG is 0.83, 0.7, 0.73 respectively. The dataset consists of a total of 13,000 tweets with 10,000 for training set (5,000 each for women and immigrant) and 3,000 for test set (1,500 each for women and immigrant). Table 1 of the work by  \citet{4_basile2019} demonstrates more information about data distribution.

\subsection{Methods}
We adopt a mixed-method study, using a combination of the building of machine learning models (RQ1 \& RQ3) and content analysis (RQ2) as our methodology.\\

\noindent\textbf{Pre-processing.} We conducted text preprocessing using standard techniques including tokenisation and removal of non-ASCII characters such as decoding emoticons\footnote{https://github.com/carpedm20/emoji}. Additionally, other pre-processing steps such as removal of punctuations and shortened URLs were performed while fine-tuning deep learning based models like DistilBERT \cite{16_Sanh2019}. We retained hashtags as these were important features of our models. \\ 

\noindent\textbf{Building of machine learning models.} We adopted state-of-the-art NLP and deep learning techniques for text classification to solve cyberbullying detection problem, and built our models using  DistilBERT \cite{16_Sanh2019}, a lighter and a faster pre-trained language model based on BERT \cite{12_Devlin2018}. 
To answer RQ1 through model comparisons, we utilised MFC and top-ranked systems in each task of HatEval challenge \cite{4_basile2019} as our baselines.To answer RQ3, we apply our pre-trained models on HatEval into other benchmark datasets related to cyberbullying. For this, we utilise three external datasets developed for SemEval Task 12 - OffensEval2020 \cite{17_zampieri2020}, SemEval Task 6 - OffensEval2019 \cite{5_zampieri2019} and Hate \& Offensive language detection by \citet{6_Davidson2017}. \\

\noindent\textbf{Content analysis.} We adopted open coding \cite{18_Corbin1990}, a qualitative content analysis technique as our method to answer RQ2 on exploring the content of misclassified tweets and categorisation them into a coding schema. 

\section{Model Description}
\subsection{Ensemble model - Task A}
\begin{figure}
    \includegraphics[width=7.5cm, height=5cm]{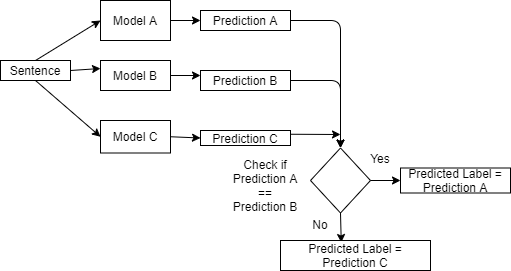}
    \caption{Deriving the final labels of Task A}
    \label{fig:taskA_model}
\end{figure}

To address the Task A we created three classification models named A, B and C (see Figure \ref{fig:taskA_model}) based on the DistilBERT model with a sequence classification head on top \cite{16_Sanh2019}. An imbalanced subset of training data where the majority class was positive was used to train model A, and an imbalanced subset of training data where the majority class was negative was used to train model B. Inspired by the approach described in \citet{20_Khoussainov}, model C was trained on a balanced subset of training data which were classified differently by the biased classifiers A and B. We fine tuned all three classifiers with a learning rate of 5e-05 for 3 epochs using a batch size of 32.Finally, we used simple voting to create an ensemble classifier combining the models A, B and C.

\subsection{Ensemble model - Task B}
Task B can be modelled as a multi-class (i.e. 5 classes) classification problem with the individual classes being (HS=0,TR=0,AG=0), (HS=1, TR=1, AG=0), (HS=1, TR=1, AG=1), (HS=1, TR=0, AG=1), and (HS=1, TR=0, AG=0) \cite{mitre}. We developed 5 binary classifiers, one for each class, using the DistilBERT model with a sequence classification head on top. Each classifier was fine tuned with a learning rate of 5e-5 and a batch size of 32 for 3 epochs. We then combined the predictions from these classifiers using probabilities to derive the final class labels. 

If only one classifier predicted a given data instance as positive, we assigned the class label of that classifier to the data instance. Whenever several classifiers predicted the positive class label for a given instance, we selected the prediction with the highest probability. If all the classifiers predicted the negative class label for a given instance, we selected the prediction with the lowest probability.

\section{Results and Discussion}
\textbf{Evaluation Metric.} To calculate the classification effectiveness, we use different metrics in each subtask. Task A uses the \textit{macro-averaged F1 score} while Task B uses \textit{Exact Match Ratio (EMR)} along with \textit{macro-averaged F1 score} \cite{4_basile2019}.

\begin{itemize}
\item \textbf{F1 Score.} The harmonic mean of precision and recall where precision is the proportion of predicted positive instances that are actually positive while recall is the proportion of actual positive instances that are predicted as positive. 
\item \textbf{Exact Matching Ratio (EMR).} Since Task B is a multi-label classification problem, EMR is calculated by combining all the dimensions (i.e. HS, TR, AG) to be predicted. The calculation of EMR is discussed in \citet{4_basile2019}.
\end{itemize}

\noindent\textbf{Baselines.} To evaluate our models (see Section 5.1), we utilise top-ranked HatEval systems and a system used by \citet{4_basile2019} as our baselines,
\begin{enumerate}
\item \textbf{Task A}. Fermi \cite{2_indurthi2019} using the SVM model with Google’s Universal Sentence Encoder \cite{19_cer2018} (refer as ‘SVM+USE’) surpassed SVC and MFC baselines of HatEval challenge.
\item \textbf{Task B}. LT3 \cite{23_LT3} ranked top in Task B.
\item \textbf{MFC baseline}. MFC is a trivial model that assigns the most frequent label in training set to all instances in the test set. 
\end{enumerate}

\subsection{Answering RQ1 - Model Evaluation}

\begin{figure}
    \includegraphics[width=8cm,height =4.5cm]{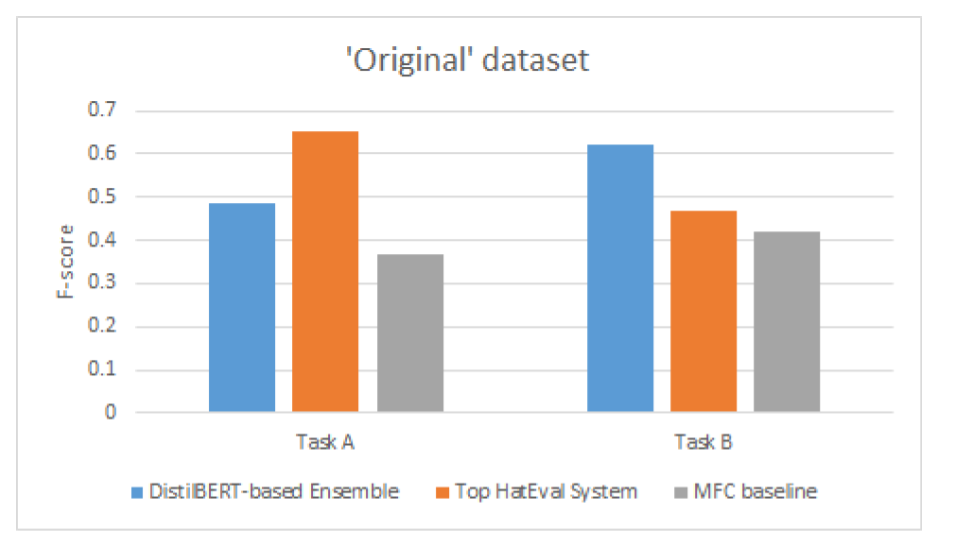}
    \caption{Model performance of Task A \& B using ‘original’ HatEval test dataset}
    \label{fig:performance_original}
\end{figure}

The performance of our ensemble model using the official HatEval test set is shown in Figure \ref{fig:performance_original}. 
The results demonstrate that our ensemble model has achieved 0.49 F1 score for Task A. In Task A, even though we outperformed MFC baseline (F1 score = 0.37), our scores did not exceed the best HatEval system - Fermi (F1 score = 0.65) \cite{2_indurthi2019}. Nevertheless, our Task A performance scores are not promising for real-world adoption.

Conversely, our ensemble model has obtained 0.62 of F1 score for task B which exceeds the best systems of HatEval Task B - LT3 \cite{23_LT3} (F1 score = 0.47) and MFC baseline (F1 score = 0.42) \cite{4_basile2019}. In Task B of HatEval, no system has been able to outperform the EMR score of MFC baseline, which achieved 0.58 of EMR (Note: \textit{Exact Matching Ratio} was the metric used for HatEval Task B evaluation). LT3 system and our ensemble model both equally achieved 0.57 of EMR which ranked us in the top place for Task B followed by MFC baseline. Since our DistilBERT-based ensemble model achieved an F1 score over 0.9 in another cyberbullying-related task (SemEval Task 12 - OffenseEval 2020) \cite{17_zampieri2020}\cite{Herath2020}, we further analysed the peculiar behaviour of model performance with HatEval challenge by unpacking the dataset.


\begin{figure*}
    \includegraphics[width=17cm, height=8cm]{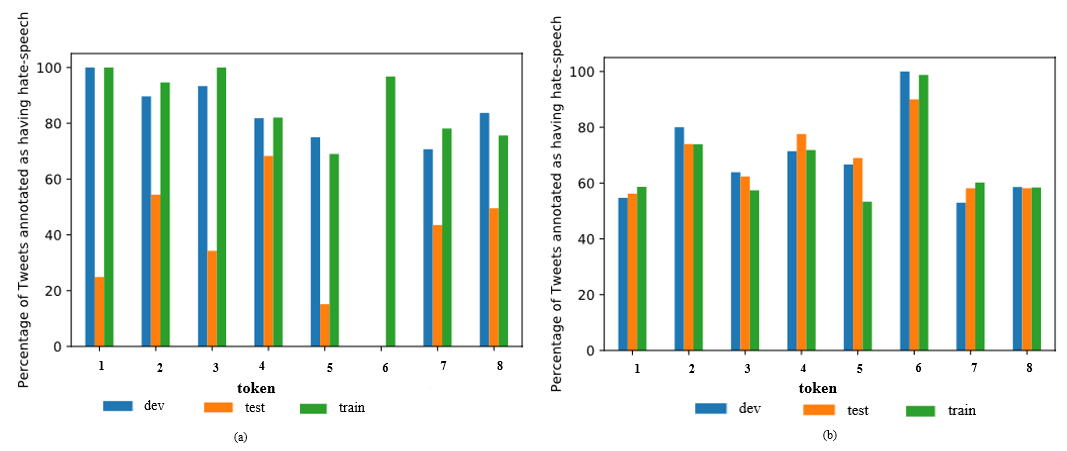}
    \caption{Variation of data across training, dev and test sets in (a) ‘original’ (b) ‘adjusted’ dataset; 1:\#buildthatwall, 2:\#buildthewall, 3:\#nodaca, 4:\#sendthemback, 5:\#stoptheinvasion, 6:\#womens**k, 7:b***h, 8:h*e}
    \label{fig:dataissue_full}
\end{figure*}

We plotted the percentages of tweets annotated as having hate speech when some common hashtags or derogatory tokens (e.g. \#buildthatwall, b***h) were found in tweets. Figure \ref{fig:dataissue_full}a) depicts the variation of data across training, dev and test sets. According to Figure \ref{fig:dataissue_full}a), it appears that training and dev set are slightly similar yet drastically different from the test set. For example, it appears that the likelihood of tweets with the token ‘\#buildthatwall (token 1)’ being annotated as having hate speech is 100\% in train and dev set, however, it is approximately 20\% in the test set.

\begin{figure}
    \includegraphics[width=8cm,height =4.5cm]{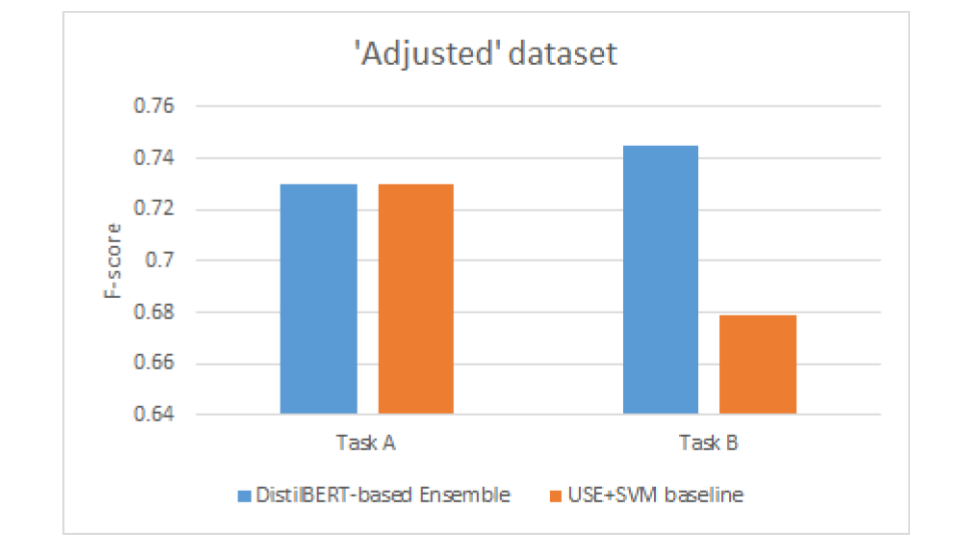}
    \caption{Model performance of Task A \& B using 'adjusted' HatEval test dataset}
    \label{fig:performance_adjusted}
\end{figure}

In order to examine whether discrepancies in the dataset had any impact on the poor performance, we merged development, training and test sets, shuffled the rows, and randomly split them again (referred to as ‘adjusted’ dataset) according to the proportions in the ‘original’ HatEval dataset (see Section 3 - 'dataset'). Figure \ref{fig:dataissue_full}b) demonstrates that there was a disparity with data distribution in the ‘original’ dataset. For example, in the ‘adjusted’ dataset, the percentage of ‘\#buildthatwall’ being annotated as having hate speech is approximate ($\sim$60\%) across train, dev and test sets. This finding led us to train our models with ‘adjusted’ dataset and fine-tuned the parameters. Figure \ref{fig:dataissue_full}b) and Figure \ref{fig:performance_adjusted} depicts the new data distribution and model performance using 'adjusted' dataset respectively.


According to Figure \ref{fig:performance_adjusted}, our ensemble models have achieved 0.73 of F1 score for Task A and 0.75 of F1 score for Task B on 'adjusted' test set. We also achieved 0.62 EMR for Task B on test set. Due to the difficulty in replicating LT3 system \cite{23_LT3} to train on 'adjusted' dataset, we obtained performance of 'SVM+USE' model \cite{2_indurthi2019} using our 'adjusted' dataset. As shown in Figure \ref{fig:performance_adjusted}, our model and baseline demonstrated equal performance in Task A. Conversely, our model outperforms 'SVM+USE' baseline by a margin of 0.06 in Task B. As mentioned in Section 4, we used 3 epochs, a batch size of 32 and a learning rate of 5e-5 to train our models. 

\begin{tcolorbox}
\textbf{RQ1:} We can automatically classify cyberbullying against women and immigrant with an F1 score of 0.73 and 0.75 in Task A (hate speech) and Task B (aggressive and targeted) respectively.
\end{tcolorbox}

The primary focus of our research is on improving the \textit{recall}, i.e. to correctly identify tweets that are cyberbullying attempts against women and immigrants as it will eventually contribute to safe cyberspace for minorities. We have achieved 0.73 and 0.76 of \textit{recall} for Task A and B respectively using 'adjusted' dataset compared to low recall of baseline systems. We are also interested in controlling true negatives, i.e. tweets that are not actually cyberbullying but are identified as positive. We exceed precision of 0.73 in both tasks using our DistilBERT-based ensemble models. Otherwise, incorrect classification of cyberbullying will have an impact on the reputation of social media platforms, particularly for freedom of speech.

\subsection{Answering RQ2 - Content Analysis of Misclassified Tweets}
To answer our RQ2, we extracted misclassified tweets (task A \& B) from our ensemble model. A content analysis method (‘open coding’) \cite{18_Corbin1990} has been adopted. The second author manually categorised 10 random misclassified tweets into three meaningful codes: gender-related issues (GEND), context-related issues (CNTX), and slangs (SLNG). After defining initial codes, two annotators (first and third author who are experienced in cyberbullying context) trialed them on a random sample of 299 misclassified tweets (population is 626 tweets), resulting in a confidence interval of 4.1 at a confidence level of 95\%. To measure the inter-annotator agreement we used the Kappa statistic.Due to the complex nature of cyberbullying phenomenon and availability of multiple codes to annotate, we failed to reach a reasonable inter-rater agreement. To overcome this, we refined our codes until we reach an agreement on a coding scheme that contained codes for all misclassified tweets in our sample. Finally, we added three additional codes: errors in original annotation (ERROR), misclassified by our model (MSCL), and not belong to any category (OTHER) when both annotators agree that original (HatEval) annotation is dubious, predicted label is incorrect, and when all other possibilities have been exhausted respectively. Table \ref{tab:RQ2_misclassified} shows the finalised set of codes along with their frequency distribution (\%).

Our results demonstrate that the lack of contextual information to resolve pronouns or user names in tweets to determine ‘gender’ (i.e. whether the target is \textit{women}) is one of the common reasons for misclassification. Based on the frequency distribution ('last column' in Table \ref{tab:RQ2_misclassified}), the most frequent category of misclassification is ‘CNTX’. Lack of contextual information is a widely raised constraint within the majority of previous works which aligns with our findings. The least frequent category of misclassification is ‘SLNG’. One possible explanation for this behaviour could be due to the dataset is extracted from an ‘adult’ group, and they are less likely to introduce new slang words compared to adolescents. Also, our results suggest that 3\% of misclassified tweets are due to ‘errors’ ($\sim$10 tweets) in the original annotations. 

Conversely, we admit that our model predicted inaccurate labels in 3\% of cases ($\sim$10 tweets). Our findings suggest that 30\% of instances belong to ‘OTHER’ category. Through manual inspection, we observed that this might be due to reasons like sarcasm, swearing with friends, abbreviations, complaints, and negations. However, the analysis reported in this paper is not comprehensive to include adequate evidence to report subcategories. 

\begin{table*}
\begin{tabular}{lllll}
\hline
\textbf{Code} & \textbf{Definition} & \textbf{Example} & \textbf{Explanation} & \textbf{(\%)}\\
\hline
GEND & \makecell{Gender-related \\ issues} & \textit{\makecell{You seem like a h*e Ok b***h? \\Did I ever deny that? Nope, Next.}} & \makecell{Misogynistic if ‘you’ \\refers to a female} & 11 \\
CNTX & \makecell{Lack of \\context} & \textit{\makecell{@user you deserve alllll the hate \\you get you h*e a*s b***h, out here \\being a damn Hypocrite you \\and cash some damn FAKES. H**s}} & \makecell{Misogynistic if ‘@user’ \\refers to a female} &44 \\
SLNG & \makecell{Issues in \\resolving slang} & \makecell{you a gay a*s b***h who seeks \\attention, STOP! I knew ever since \\you gonna switch up on me... I guess \\you did F***ING SNAKE A*S H*E!} & \makecell{Non-misogynistic if \\‘gay a*s b***h’ \\slang is resolved} & 9 \\
ERROR & \makecell{Issues in \\original annotation} & \makecell{@user It means $<$religion$>$ will show \\ them how to rape/abuse women 24/7!} & \makecell{Targeted to immigrants} & 3 \\
MSCL & \makecell{Misclassified \\by our model} & \makecell{Europe is being invaded by third world \\"refugees" Continue to Pray for them} & \makecell{Targeted refugees} & 3 \\
OTHER & \makecell{Not belong to \\any category} & \makecell{REFUGEES NOT WELCOME} & \makecell{Targeted refugees \\if negation is recognised} &30 \\
\hline
\end{tabular}
\caption{Coding reference of misclassified tweets.}
\label{tab:RQ2_misclassified}
\end{table*}

\begin{tcolorbox}
\textbf{RQ2:} Misclassified tweets can be categorised into six types, with the context-related issues (‘CNTX’) being the most frequent reason for misclassification, followed by issues to resolve gender (‘GEND’) and slang (‘SLNG’).
\end{tcolorbox}

\subsection{Answering RQ3 - Cross-task Evaluation}
To answer our RQ3 about the generalisability of our models on different cyberbullying-related tasks, we applied and tested our pre-trained ensemble model in other three tasks, 1) SemEval 2020 - Task 12 (OffensEval2020) \cite{17_zampieri2020}, 2) SemEval 2019 - Task 6 (OffensEval2019) \cite{5_zampieri2019}, and 3) Hate \& Offensive language (refer as ‘Hate \& Offense’) dataset \cite{6_Davidson2017}. 
\begin{enumerate}
    \item \textbf{OffensEval datasets \cite{5_zampieri2019,17_zampieri2020}}. These datasets include three subtasks to determine whether a tweet expresses cyberbullying based on whether it is, 1) offensive or not, 2) targeted or not, and 3) if targeted, whether it is toward an individual, group, or other. 
    
    \item \textbf{Hate \& Offense dataset \cite{6_Davidson2017}.} This dataset also has three subtasks to determine whether a tweet include hate and offensive language based on, 1) hate speech or not, 2) offensive but not hate speech, and 3) neither offensive nor hate speech.
\end{enumerate}

The tasks of these two datasets were different from HatEval challenge except the first subtask to determine hate (or offensive) language. Therefore, we report the results of cross-domain validation using Task A (i.e. hate speech or not) only.We extracted a random sample of 2,971 tweets from each dataset to align with the test size of our original Task A when the official test set was unavailable publicly. \ref{tab:RQ3performance} shows the outcome using our pre-trained ensemble model (Task A). 

\begin{table}
\begin{tabular}{llllll}
\hline
\textbf{Dataset} & \textbf{\makecell{Sample \\size}} & \textbf{Acc.} & \textbf{P} & \textbf{R} & \textbf{F1} \\
\hline
OffensEval2020 & 3887 & 0.74 & 0.72 & 0.74 & 0.68 \\
OffensEval2019 & 860 & 0.68 & 0.66 & 0.68 & 0.67 \\
Hate \& Offense & 2971 & 0.70 & 0.74 & 0.70 & 0.69 \\
\hline
\end{tabular}
\caption{\label{table1_performance}Performance (weighted average) of our pre-trained Task A model on other cyberbullying-related tasks; Acc.:Accuracy, P:Precision, R:Recall.
}
\label{tab:RQ3performance}
\end{table}

Current state of the art models have reportedly achieved F1 scores of 0.82, 0.92 and 0.90 for OffensEval2019, OffensEval2020 and Hate \& Offense datasets respectively \cite{5_zampieri2019,17_zampieri2020}\cite{6_Davidson2017}. According to Table \ref{tab:RQ3performance},  we have achieved a satisfactory performance with approximately 0.7 of accuracy/F1 score for all task pairs (i.e. training on HatEval dataset and testing on another dataset).These results suggest that our pre-trained ensemble model on HatEval is generalised reasonably well (Accuracy/F1 score $\sim$0.7) when classifying \textit{hate speech} irrespective of the context (e.g. misogyny etc.). Due to the misalignment between datasets, we did not apply our models to other tasks of external datasets. 

\begin{tcolorbox}
\textbf{RQ3:} Our pre-trained models from HatEval dataset can automatically classify hate speech in other benchmarking datasets with a reasonable accuracy ($\sim$0.7). 
\end{tcolorbox}

\section{Discussion}
The ultimate goal of our work is to advance the technology to detect and classify cyberbullying using state-of-the-art NLP techniques, with the long-term aim of enabling social media as a safe space for all users. We developed DistilBERT-based ensemble model per task as a basis to answer our RQ1. With an initial poor performance using a test set of 'original' HatEval dataset, we suggest developing a novel version of the original dataset (i.e. 'adjusted') through merging, shuffling and splitting. The \textit{'adjusted'} dataset contributed to better performance of F1 score of 0.73 and 0.74 for Task A and B respectively. 

The six categories of misclassified tweets that emerged from our qualitative analysis (RQ2) build a point of reference for the content of such misclassifications. This initial categories can help researchers to understand the grounds to improve automated cyberbullying classification. Also, the categories identified through this research can serve as a guide which could extend as a conceptual framework for future qualitative and quantitative cyberbullying research. Additionally, the categories along with the frequencies that we report in this work provide implications for researchers to collect, annotate, and revise their datasets that could minimise the likelihood of misclassification produced by machine learning models including providing additional contextual information about data. Conversely, this raises new research questions on whether we could improve the performance of machine learning models further without relying on demographic data such as \textit{gender} and data on language evolution such as \textit{out-of-vocabulary slang} and \textit{abbreviations}.

The findings from our RQ3 on generalisability of pre-trained models on other cyberbullying-related tasks demonstrated reasonable accuracy ($\sim$0.7). A possible explanation of not achieving more could be due to pre-trained models might biased within women and immigrant context (e.g. specific hashtags, misogyny) and not be the best option for classifying ‘general’ offense-related tasks. As a solution, future models could augment data from general as well as specific datasets (e.g., racial \cite{14_davidson2019}, gendered \cite{24_gendered}).

In addition to the lack of contextual information that limits our model improvement further, this research is subject to implicit bias of annotators when judging categories to answer RQ2. As a solution, our future work will incorporate a semi-automated approach for misclassification annotation by reusing readily available lexical resources like MRC psycholinguistic database \cite{21_Colheart}, LIWC \cite{22_ennebaker2001} to obtain initial codes and employ at least three annotators to refine the codes. Furthermore, our 'adjusted' dataset may not provide a robust solution in terms of replicability. Therefore, we intend to create a couple of 'adjusted' datasets and report the average of performance in our future works. We also share our current 'adjusted' dataset to enable replication of experiments.

In summary, we propose that future cyberbullying classification models need to concentrate on incorporating state-of-the-art solution to common NLP problems like language evolution, sarcasm detection, and pronoun resolution. Additionally, future research should also focus on advancing the prediction of demographic information such as gender, age, and personality from data within an ethical framework without reidentifying Twitter profiles.

\section{Conclusions}
Due to massive participation in social media, manual moderation of cyberbullying is an extremely labour-intensive task which leads to delay in taking action against bullies while protecting victims. Accordingly, automated classification of cyberbullying emerged and remains as a challenging NLP task. This research contributes to develop machine learning models for cyberbullying classification. Through a qualitative content analysis, we also contributed to develop a coding schema to deepen the understanding of misclassifications produced by models, enabling future researchers to minimise the impact of data for poor model performance. When social media platforms are equipped with effective cyberbullying detection models, victimised communities will be able to discuss their concerns openly, without harassment.

\bibliography{acl2020}
\bibliographystyle{acl_natbib}

\end{document}